\documentclass{article}

\usepackage{microtype}
\usepackage{graphicx}
\usepackage{booktabs} 

\usepackage{hyperref}

\usepackage{subcaption}

\usepackage{amsfonts}
\usepackage{tikz}
\newcommand*\circled[1]{\tikz[baseline=(char.base)]{
		\node[shape=circle,draw,inner sep=0.5pt] (char) {#1};}}
\usepackage{times}
\usepackage{latexsym}
\usepackage{amsmath} 
\usepackage{amsthm}
\usepackage{amssymb}
\usepackage{latexsym} 
\usepackage{pgfplots}
\usepackage{tikz}
\usepackage{multirow}
\usepackage{arydshln}
\usepackage{times}

\include{plots}
\definecolor{bblue}{HTML}{4F81BD}
\definecolor{rred}{HTML}{C0504D}
\definecolor{ggreen}{HTML}{9BBB59}
\definecolor{ppurple}{HTML}{9F4C7C}
\definecolor{Dark scarlet}{HTML}{560319}
\definecolor{Forest green}{HTML}{1E4D2B}

\usepackage[accepted]{icml2020}

\icmltitlerunning{Neural Topic Modeling with Continual Lifelong Learning}

\begin{document}
	
	\twocolumn[
	\icmltitle{
		Neural Topic Modeling with Continual Lifelong Learning
	}
	
	
	
	
	\begin{icmlauthorlist}
		\icmlauthor{Pankaj Gupta}{goo}
		\icmlauthor{Yatin Chaudhary}{goo,to}
		\icmlauthor{Thomas Runkler}{goo}
		\icmlauthor{Hinrich Sch\"utze}{to}
	\end{icmlauthorlist}
	
	\icmlaffiliation{goo}{Corporate Technology, Siemens AG Munich, Germany}
	\icmlaffiliation{to}{CIS, University of Munich (LMU) Munich, Germany}
	
	\icmlcorrespondingauthor{Pankaj Gupta}{pankaj.gupta@drimco.net}
	
	\icmlkeywords{Machine Learning, ICML}
	
	\vskip 0.3in
	]
	
	
	
	\printAffiliationsAndNotice{}  
	
	\begin{abstract}
		Lifelong learning has recently attracted attention in building machine learning systems that continually accumulate and transfer knowledge to help future learning. Unsupervised topic modeling has been popularly used to discover topics from document collections. However, the application of topic modeling is challenging due to data sparsity, e.g., in a small collection of (short) documents and thus, generate incoherent topics and sub-optimal document representations. To address the problem, we propose a lifelong learning framework for neural topic modeling that can continuously process streams of document collections, accumulate topics and guide future topic modeling tasks by knowledge transfer from several sources to better deal with the sparse data. In the lifelong process, we particularly investigate jointly: (1) sharing generative homologies (latent topics) over lifetime to transfer prior knowledge, and (2) minimizing catastrophic forgetting to retain the past learning via novel selective data augmentation, co-training and topic regularization approaches. Given a stream of document collections, we apply the proposed Lifelong Neural Topic Modeling (LNTM) framework in modeling three sparse document collections as future tasks and demonstrate improved performance quantified by perplexity, topic coherence and information retrieval task.
		Code: \url{https://github.com/pgcool/Lifelong-Neural-Topic-Modeling}
	\end{abstract}
	
	\section{Introduction}
	{\it Unsupervised topic models}, such as LDA \cite{DBLP:journals/jmlr/BleiNJ03}, RSM \cite{DBLP:conf/nips/SalakhutdinovH09}, DocNADE \cite{DBLP:journals/jmlr/LaulyZAL17}, NVDM \cite{SrivastavaSutton}, etc. 
	have been popularly used to discover topics from large document collections.  
	However in sparse data settings, the application of topic modeling is challenging due to limited context in a small document collection or short documents (e.g., tweets, headlines, etc.) 
	and the topic models produce incoherent topics. 
	To deal with this problem, there have been several attempts \cite{DBLP:conf/nips/PettersonSCBN10,P15-1077,DBLP:journals/tacl/NguyenBDJ15,pankajgupta:2019iDocNADEe} 
	that introduce prior knowledge such as pre-trained word embeddings \cite{D14-1162} to guide meaningful learning. 
	\begin{figure}[t]
		\begin{center}
			\includegraphics[scale=0.61]{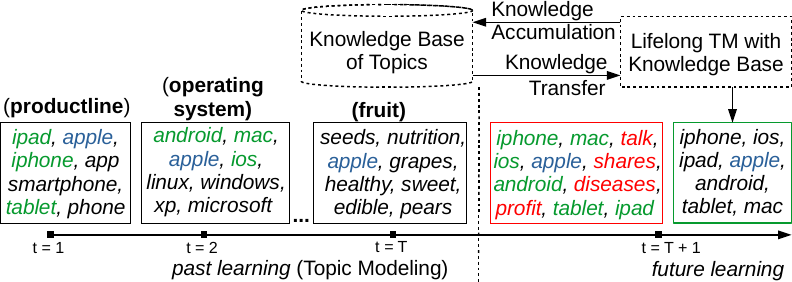}
		\end{center}
		\vskip -0.2in
		\caption{Motivation for Lifelong Topic Modeling}
		\label{fig:lifelongNTMmotvation}
		\vskip -0.2in
	\end{figure}
	
	{\it Lifelong Machine Learning} (LML) \cite{thrun1995lifelong,DBLP:conf/aaai/MitchellCHTBCMG15,hassabis2017neuroscience,DBLP:journals/nn/ParisiKPKW19} 
	has recently attracted attention in building adaptive computational systems that can continually acquire, retain and transfer knowledge over life time 
	when exposed to modeling continuous streams of information.  
	In contrast, the traditional machine learning is based on isolated learning i.e., a one-shot task learning (OTL) using a single dataset and thus, 
	lacks ability to continually learn from incrementally available heterogeneous data.  
	The application of LML framework has shown potential for supervised natural language processing (NLP) tasks \cite{chen-liu-2016-lifelong} 
	such as in sentiment analysis \cite{DBLP:conf/acl/ChenM015}, relation extraction \cite{DBLP:conf/naacl/WangXYGCW19}, text classification \cite{DBLP:conf/nips/dAutumeRKY19}, etc. 
	Existing works in topic modeling are either based on the OTL approach or transfer learning \cite{DBLP:conf/icml/Chen014} using stationary batches of training data and prior knowledge 
	without accounting for streams of document collections. 
	The unsupervised document (neural) topic modeling still remains unexplored regarding lifelong learning. 
	
	In this work, we explore unsupervised document (neural) topic modeling within a continual lifelong learning paradigm to enable knowledge-augmented topic learning over lifetime. 
	We show that {\it Lifelong Neural Topic Modeling} (LNTM) is capable of mining and retaining prior knowledge (topics) from streams of large document collections, 
	and particularly guiding topic modeling on sparse datasets using accumulated knowledge of several domains over lifespan.  
	For example in Figure \ref{fig:lifelongNTMmotvation}, we have a stream of coherent topics associated with {\it apple} extracted from a stream of large document collections over time $t \in [1, T]$ (i.e., past learning). 
	Observe that the word {\it apple} is topically contextualized by several domains, i.e., \texttt{productline}, \texttt{operating system} and \texttt{fruit} at tasks $t=1$, $t=2$ and $t=T$, respectively.  
	For the future task $T+1$ on a small document collection, the topic (red box) produced without LNTM is incoherent, 
	containing some irrelevant words (marked in red) from various topics.  
	Given a sufficient overlap (marked in green) in the past and future topic words, we aim to help topic modeling for the future task $T+1$ such that the topic (red box) becomes semantically coherent (green box), 
	leading to generate an improved document representation.   
	
	Therefore, the goal of LNTM is to 
	(1) detect topic overlap in prior topics $t \in [1, T]$ of the knowledge base (KB) and topics of future task $T+1$, 
	(2) positively transfer prior topic information in modeling future task,  
	(3) retain or minimize forgetting of prior topic knowledge, and 
	(4) continually accumulate topics in KB over life time.  
	In this work, we particularly focus on addressing the challenge: {\it how to simultaneously mine relevant knowledge from prior topics,  transfer mined topical knowledge and also retain prior topic information under domain shifts over lifespan?}  
	
	{\bf Contributions}: We present a novel lifelong neural topic modeling framework that learns topics for a future task with proposed approaches of: 
	(1) {\it Topic Regularization} that enables topical knowledge transfer from several domains and prevents catastrophic forgetting in the past topics, 
	(2) {\it Word-embedding Guided Topic Learning} that introduces prior multi-domain knowledge encoded in word-embeddings, and 
	(3) {\it Selective-data Augmentation Learning} that identifies relevant documents from historical collections, learns topics simultaneously with a future task and controls forgetting due to selective data replay. We apply the proposed framework in modeling three sparse (future task) and four large (past tasks) document collections in  sequence. Intensive experimental results show improved topic modeling on future task while retaining past learning, quantified by information retrieval, topic coherence and generalization capabilities. 
	
	\begin{table}[t]
		\vskip -0.1in
		\caption{Description of the notations used in this work}
		\label{tab:notations}
		\begin{center}
			\begin{small}
				\renewcommand*{\arraystretch}{1.2}
				\resizebox{.49\textwidth}{!}{
					\begin{tabular}{c|l}
						\toprule
						\multicolumn{1}{c|}{\bf Notation} & \multicolumn{1}{c}{\bf Description} \\ 
						\midrule
						LNTM    & Lifelong Neural Topic Modeling								\\ 
						EmbTF  &  Word Embedding based transfer  					    \\ 
						TR & Topic Regularization  		 									\\
						SAL    & Selective-data Augmentation Learning					  				  					\\
						\texttt{TopicPool}		& Pool of accumulated topics 				  \\
						\texttt{WordPool}		& Pool of accumulated word embeddings 	\\
						$\Omega^t$		&  A document collection at time/task $t$ 				\\
						$(T+1)$			& Future task 								\\ 
						$\{1,...,T\}$		& Past tasks 									\\ 
						${\bf Z}^t \in \mathbb{R}^{H \times K}$		& Topic Embedding matrix	for task $t$					\\ 
						${\bf E}^t \in \mathbb{R}^{E \times K}$		& Word Embedding matrix for task $t$ 					\\ 
						${\bf \Theta}$	& LNTM parameters 				\\
						${\bf \Phi}$	       & LNTM hyper-parameters 				\\  
						$ \lambda^{t}_{EmbTF}$ 	 & Degree of relevance of ${\bf E}^{t}\in$ \texttt{WordPool} for $(T+1)$ 	 \\
						$ \lambda^{t}_{TR}$  &	Degree of topic imitation/forgetting of ${\bf Z}^{t}$ by ${\bf Z}^{T+1}$		 \\ 
						$ \lambda^{t}_{SAL}$  &	Degree of domain-overlap in ${\Omega}^{t}$ and ${\Omega}^{T+1}$		 \\ 
						${\bf A}^{t} \in \mathbb{R}^{H \times H}$    & Topic-alignment in ${\bf Z}^{t}$ and $Z^{T+1}$ 		 \\ 
						$ K, D$  & 	Vocabulary size, document size	 \\ 
						$E$,  $H$ & Word embedding dimension, \#topics  \\ 
						${\bf b} \in \mathbb{R}^{K}$   & Visible (input) bias vector   \\
						${\bf c} \in \mathbb{R}^{H}$    &  Hidden (input) bias vector 	\\ 
						${\bf v}$  & An input document (visible units)	   	\\
						$\mathcal{L}^{t}$ & Loss (negative log-likelihood) for task $t$	 \\
						${\bf W} \in \mathbb{R}^{H \times K}$   & Encoding matrix of DocNADE for task $(T+1)$	  \\ 
						${\bf U} \in \mathbb{R}^{K \times H}$   & Decoding matrix of DocNADE for task $(T+1)$ \\
						\bottomrule
				\end{tabular}}
			\end{small}
		\end{center}
		\vskip -0.2in
	\end{table}
	
	\begin{figure*}[t]
		\begin{center}
			\centerline{\includegraphics[scale=0.7]{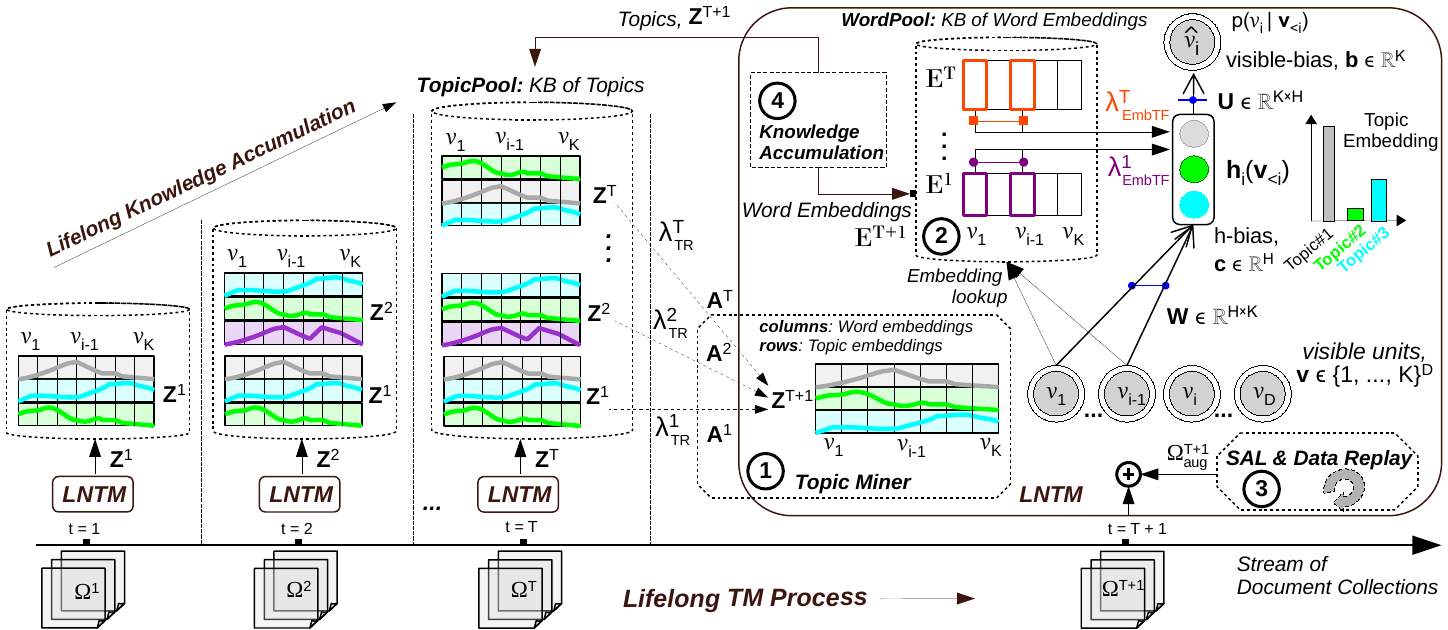}}
			\caption{An illustration of the proposed Lifelong Neural Topic Modeling (LNTM) framework over s stream of document collections}
			\label{fig:lifelongNTM}
			\vskip -0.1in
		\end{center}
		\vskip -0.2in
	\end{figure*}
	
	\section{Methodology: Lifelong Topic Modeling}
	In following section, we describe our contributions in building Lifelong Neural Topic Modeling framework including: 
	topic extraction, knowledge mining, retention, transfer and accumulation. {\it See Table \ref{tab:notations} for the description of notations}. 
	
	Consider a stream of document collections $\bf S$ = \{$\Omega^{1}$, $\Omega^{2}$,..., $\Omega^{T}$, $\Omega^{T+1}$\} over lifetime $t \in [1,...,T, T+1]$, where $\Omega^{T+1}$ is used to perform future learning. 
	During the lifelong learning, we sequentially iterate over $\bf S$ and essentially analyze a document collection $\Omega^{t} \in \bf S$ using a novel topic modeling framework that can leverage and 
	retain prior knowledge extracted from each of the lifelong steps $\{1,..., t-1\}$.   
	
	\subsection{Topic Learning via Neural Topic Model}\label{sec:docnade}
	Within the OTL framework, an unsupervised neural-network based topic model named as Document Neural Autoregressive Distribution Estimation (DocNADE) \cite{DBLP:conf/nips/LarochelleL12,DBLP:journals/jmlr/LaulyZAL17} 
	has shown to outperform existing topic models based on LDA \cite{DBLP:journals/jmlr/BleiNJ03,SrivastavaSutton} 
	or neural networks such as Replicated Softmax (RSM) \cite{DBLP:conf/nips/SalakhutdinovH09}, Autoencoders \cite{DBLP:journals/jmlr/LaulyZAL17}, NVDM \cite{DBLP:conf/icml/MiaoYB16} etc. 
	Additionally, \citeauthor{pankajgupta:2019iDocNADEe} \yrcite{pankajgupta:2019iDocNADEe} have recently demonstrated competitiveness of DocNADE in transfer learning settings. 
	Thus, we adopt DocNADE as the backbone in discovering topics and building lifelong topic learning framework.
	
	{\bf DocNADE Formulation:} For a document (observation vector) ${\bf v} \in \Omega$ of size $D$ such that ${\bf v} = (v_1, ...v_D)$, each word index $v_i$  takes a value in vocabulary $\{1, ..., K\}$ of size $K$. 
	Inspired by NADE \cite{DBLP:journals/jmlr/LarochelleM11} and RSM \cite{DBLP:conf/nips/SalakhutdinovH09} generative modeling architectures, 
	DocNADE computes the joint probability distribution $p({\bf v} ; {\bf \Theta}) = \prod_{i=1}^{D} p(v_i | {\bf v}_{<i}; {\bf \Theta})$ of 
	words in the document ${\bf v}$ by factorizing it as a product of conditional distributions $p(v_i | {\bf v}_{<i};{\bf \Theta})$, 
	where each conditional is efficiently modeled via a feed-forward neural network using proceeding word ${\bf v}_{<i}$ in the sequence. 
	
	Following reconstruction principle, the DocNADE computes a hidden vector ${\bf h}_i ({\bf v}_{<i})$ at each autoregressive step:
	\begin{align*}\label{eq:DocNADEconditionals}
		\begin{split}
			{\bf h}_i({\bf v}_{<i})    =  g ({\bf c} + & \sum_{q<i} {\bf W}_{:, v_q}) \ \mbox{and} \  g=\{sigmoid, tanh\}\\ 
			p (v_i = w | {\bf v}_{<i}; {\bf \Theta})   & = \frac{\exp (b_w + {\bf U}_{w,:} {\bf h}_i ({\bf v}_{<i}))}{\sum_{w'} \exp (b_{w'} + {\bf U}_{w',:} {\bf h}_i ({\bf v}_{<i}))}
		\end{split}
	\end{align*}
	for each $i \in \{1,...D\}$, where 
	${\bf v}_{<i} \in \{v_1, ...,v_{i-1}\}$ is a sub-vector consisting of all $v_q$ such that $q < i$. 
	${\bf \Theta}$ is a collection of parameters including weight matrices $\{{\bf W} \in \mathbb{R}^{H \times K}$, ${\bf U} \in \mathbb{R}^{K \times H}$\} and biases $\{{\bf c} \in \mathbb{R}^H$, ${\bf b} \in \mathbb{R}^K\}$.  
	$H$ and $K$ are the number of hidden units (topics) and vocabulary size. 
	
	Figure \ref{fig:lifelongNTM} (rightmost; without components \circled{1}, \circled{2}, \circled{3} and \circled{4}) illustrates the DocNADE architecture,  
	computing the probability ${\hat{v}}_i = p({\bf v}_i | {\bf v}_{<i}; {\bf \Theta})$ of the $i^{th}$ word $v_i$  conditioned on position dependent hidden layer ${\bf h}_i({\bf v}_{<i})$. 
	The parameter $\bf W$ is shared in the feed-forward networks and ${\bf h}_i$ encodes topic proportion for the document ${\bf v}$. 
	
	Algorithm \ref{algo:lifelonglearning} (lines \#1-4) and {\small TOPIC-LEARNING} utility (algorithm \ref{algo:lifelongfunctions}) describe the computation of objective function: 
	negative log-likelihood $\mathcal{L}({\bf v}; {\bf \Theta})$ that is minimized using stochastic gradient descent. 
	In terms of model complexity, computing ${\bf h}_i ({\bf v}_{<i})$ is efficient (linear complexity)
	due to NADE \cite{DBLP:journals/jmlr/LarochelleM11} architecture that leverages the pre-activation ${\bf a}_{i-1}$ of $(i-1)$th
	step in computing ${\bf a}_i$. 
	The complexity of computing all hidden layers ${\bf h}_i ({\bf v}_{<i})$ is in $O(DH)$ and all $p({\bf v}_i | {\bf v}_{<i}; {\bf \Theta})$ in $O(KDH)$ for $D$ words in the document ${\bf v}$. 
	Thus, the total {\it complexity} of computing the joint distribution $p({\bf v})$ is in $O(DH + KDH)$. 
	
	Importantly, the topic-word matrix ${\bf W} \in \mathbb{R}^{H \times K}$ has a property that  
	the row-vector ${\bf W}_{j, :}$ encodes $j$th topic (distribution over vocabulary words), i.e., topic-embedding   
	whereas the column-vector ${\bf W}_{:, v_i}$ corresponds to embedding of the word $v_i$, i.e., word-embedding. 
	We leverage this property to introduce prior knowledge via topic and word embeddings during lifelong learning. 
	Additionally, we accumulate all topic and word embeddings in \texttt{TopicPool} and \texttt{WordPool}, respectively learned over lifetime. 
	
	\subsection{Lifelong Learning in Neural Topic Modeling}
	Given the prior knowledge (\texttt{TopicPool} and \texttt{WordPool}), a stream of document collections ${\bf S}$ and a new (future) topic learning task on document collection $\Omega^{T+1}$, 
	the proposed LNTM framework operates in two phases: 
	
	{\bf Phase 1: Joint Topic Mining, Transfer and Retention:} 
	The task of topic modeling with lifelong learning capabilities is prone to three main challenges:  
	(a) mining prior knowledge relevant for the future task $T+1$, 
	(b) learning with prior knowledge, and 
	(c) minimizing catastrophic forgetting, i.e., retaining of prior knowledge.  
	Here, the prior knowledge refers to topic and word embeddings extracted from the historical tasks $\{1, ...T\}$. 
	In modeling a future task $T+1$, we address the above challenges by jointly mining, transferring and retaining prior knowledge.        
	Algorithms \ref{algo:lifelonglearning} and \ref{algo:lifelongfunctions} demonstrate the following three {\bf approaches} within lifelong neural topic modeling, \texttt{LNTM}= \{\texttt{TR}, \texttt{EmbTF}, \texttt{SAL}\}: 
	
	{\bf \circled{1} Topic Regularization with} \texttt{TopicPool} (\texttt{TR}): 
	To address the learning without forgetting, several works \cite{DBLP:journals/corr/JungJJK16,kirkpatrick2017overcoming,DBLP:conf/icml/ZenkePG17,DBLP:journals/pami/LiH18a} 
	have investigated regularization approaches in building LML systems that constrain the updates of neural weights in modeling the future task (i.e., $T+1$)  
	such that catastrophic forgetting with all the previously learned tasks is minimized.  These existing works majorly focus on building LML systems dealing with computer vision tasks mostly in supervised fashion. 
	Lifelong topic and document representation learning in unsupervised fashion has received considerable less attention. 
	Thus inspired by the regularization strategies, we regularize topics of the past and future tasks in a way that not only minimizes 
	forgetting of prior topics but also maximizes topical knowledge transfer for the future task (i.e., unsupervised topic modeling). 
	
	Given a pool of prior topics, i.e.,  \texttt{TopicPool} built by accumulating topics from each of the past tasks, 
	we perform topic mining for the future task $T+1$ using DocNADE.    
	In doing so, 
	the topic learning ${\bf Z}^{T+1}$ on document collection $\Omega^{T+1}$ is guided 
	by all the past topics $[{\bf Z}^{1},...{\bf Z}^{T}] \in $ \texttt{TopicPool}, building a {\it Topic Miner} that consists of:
	
	(a) {\it Topic Extractor}: A topic is essentially a distribution over vocabulary that explains thematic structures in the document collection.  
	In modeling a stream of document collections, the vocabulary size may not be same in tasks over lifetime and thus 
	topic analogy (e.g., shifts, overlap, etc.) requires common vocabulary words in the participating topics. 
	Illustrated in Figure \ref{fig:lifelongNTM}, each latent topic vector in ${\bf Z}^{T+1}$ (marked by \circled{1}) of the future task $T+1$ 
	encodes a distribution over words appearing in the past tasks, e.g., ${\bf Z}^{T} \in$ \texttt{TopicPool}.     
	As discussed in section \ref{sec:docnade}, the topics ${\bf Z}^{T+1}$ can be obtained from the row-vectors of ${\bf W} \in {\bf \Theta}^{T+1}$ 
	by masking all its column-vectors $v_i$ not in the past.
	
	(b) {\it Topic Regularizer}: Given \texttt{TopicPool}, we model the future task by introducing an additional topic-regularization term $\Delta_{TR}$
	in its objective function ${\mathcal{L}(\Omega^{T+1}; {\bf \Theta}^{T+1})}$:
	\begin{equation*}
		\Delta_{TR} =  \sum_{t=1}^{T}  \lambda_{TR}^t  (\underbrace{||  {\bf Z}^{t} - {\bf A}^{t} {\bf Z}^{T+1}  ||_2^2}_{topic-imitation}  +  \underbrace{|| {\bf U}^{t} - {\bf P}^{t} {\bf U}||_2^2}_{decoder-proximity})
	\end{equation*}
	$\qquad \mathcal{L}(\Omega^{T+1} ; {\bf \Theta}^{T+1})$ =$\sum_{{\bf v} \in \Omega^{T+1}}   \mathcal{L}({\bf v}  ; {\bf \Theta}^{T+1})  + \Delta_{TR}$ 
	
	such that the first term ({\it topic-imitation}) allows controlled knowledge transfer by inheriting relevant topic(s) in ${\bf Z}^{T+1}$ from \texttt{TopicPool}, 
	accounting for domain-shifts via a topic-alignment matrix ${\bf A}^{t} \in \mathbb{R}^{H \times H}$ for every prior task.  
	Moreover, the two terms together preserve the prior learning with encoder and decoder proximity, respectively due to a quadratic penalty on the selective 
	difference between the parameters for the past and future topic modeling tasks, such that the parameters ${\bf \Theta}^{T+1}$ 
	also retain representation capabilities for the document collections in the past, e.g.,   
	$\mathcal{L}(\Omega^{t}; {\bf \Theta}^{t}) \sim \mathcal{L}(\Omega^{t}; {\bf \Theta}^{T+1})$. 
	Here,  $\lambda_{TR}^t$ is per-task regularization strength that controls the degree of topic-imitation and forgetting of prior learning $t$ by the future task $T+1$.   
	$({\bf Z}^{t}, {\bf U}^{t}) \in {\bf \Theta}^{t}$ are parameters at the end of the past task $t$. 
	
	Figure \ref{fig:lifelongNTM} ({\it Topic Miner} component \circled{1}) and Algorithm \ref{algo:lifelonglearning} (lines \#10-14) demonstrate the \texttt{TR} approach in LNTM framework.  
	The topic regularization $\Delta_{TR}$ approach enables jointly mining, transferring and retaining prior topics when learning future topics continually over lifetime. 
	\begin{algorithm}[t]
		\caption{{\small Lifelong Neural Topic Modeling using DocNADE}}
		\label{algo:lifelonglearning}
		\begin{center}
			\begin{small}
				\begin{algorithmic}[1]
					\INPUT Sequence of document collections $\{\Omega^{1},...\Omega^{T},...\Omega^{T+1}\}$
					\INPUT Past learning: \{${\bf \Theta}^{1},...,{\bf \Theta}^{T}$\}
					\INPUT \texttt{TopicPool}: $\{{\bf Z}^{1}, ..., {\bf Z}^{T}\}$
					\INPUT \texttt{WordPool}:  $\{{\bf E}^{1}, ..., {\bf E}^{T}\}$
					\PARAMETERS ${\bf \Theta}^{T+1}$ = $\{{\bf b}, {\bf c}, {\bf W}, {\bf U},  {\bf A}^{1},...,  {\bf A}^{T}, {\bf P}^{1},...,  {\bf P}^{T}\}$
					\HYPERPARAMETERS ${\bf \Phi}^{T+1}$ = $\{H, \lambda_{LNTM}^{1},..., \lambda_{LNTM}^{T}$\} 
					\STATE {\bf Neural Topic Modeling}:
					\STATE \texttt{LNTM} = \{\}
					\STATE Train a topic model and get PPL on test set $\Omega^{T+1}_{test}$:
					\STATE $\quad$ $\mbox{PPL}^{T+1}, {\bf \Theta}^{T+1} \gets$  topic-learning($\Omega^{T+1}, {\bf \Theta}^{T+1}$)
					\item[] 
					\STATE {\bf Lifelong Neural Topic Modeling (LNTM) framework}:
					\STATE \texttt{LNTM} = \texttt{\{EmbTF, TR, SAL\}} 
					\STATE For a document ${\bf v} \in \Omega^{T+1}$: 
					\STATE Compute loss (negative log-likelihood): 
					\STATE $\quad$ $\mathcal{L}({\bf v}  | {\bf \Theta}^{T+1}) \gets$  compute-NLL(${\bf v}, {\bf \Theta}^{T+1}$, \texttt{LNTM})
					\IF {\texttt{TR} in  \texttt{LNTM}}
					\STATE Jointly minimize-forgetting and learn with \texttt{TopicPool}: 
					\STATE $\Delta_{TR} \gets  \sum_{t=1}^{T} \lambda_{TR}^t \  (||  {\bf Z}^{t} - {\bf A}^{t} {\bf Z}^{T+1}  ||_2^2  +   || {\bf U}^{t} - {\bf P}^{t} {\bf U}||_2^2)$
					\STATE $\mathcal{L}({\bf v} ; {\bf \Theta}^{T+1}) \gets \mathcal{L}({\bf v}  ; {\bf \Theta}^{T+1})  + \Delta_{TR} $
					\ENDIF
					\IF {\texttt{SAL} in  \texttt{LNTM}} 
					\STATE Detect domain-overlap and select relevant historical documents from [$\Omega^{1},...,\Omega^{T}$] for augmentation at task (T+1):
					\STATE $\Omega_{aug}^{T+1} \gets$  distill-documents(${\bf \Theta}^{T+1}$, $\mbox{PPL}^{T+1}$, [$\Omega^{1},..., \Omega^{T}$])
					\STATE Perform augmented learning (co-training) with $\Omega_{aug}^{T+1}$: 
					\STATE $\Delta_{SAL} \gets  \sum_{({\bf v}^{t}, t) \in \Omega^{T+1}_{aug}} \lambda_{SAL}^t \ \mathcal{L}( {\bf v}^{t} ;{\bf \Theta}^{T+1})$
					\STATE $\mathcal{L}({\bf v};{\bf \Theta}^{T+1}) \gets \mathcal{L}({\bf v};{\bf \Theta}^{T+1})+\Delta_{SAL}$
					\ENDIF  
					\STATE Minimize $\mathcal{L}({\bf v}  ; {\bf \Theta}^{T+1})$ using stochastic gradient-descent
					\STATE {\bf Knowledge Accumulation}: 
					\STATE \texttt{TopicPool} $\gets$ accumulate-topics(${\bf \Theta}^{T+1}$)
					\STATE \texttt{WordPool} $\gets$ accumulate-word-embeddings(${\bf \Theta}^{T+1}$)
				\end{algorithmic}
			\end{small} 
		\end{center}
	\end{algorithm}
	
	{\bf \circled{2} Transfer Learning with} \texttt{WordPool} (\texttt{EmbTF}):
	Beyond topical knowledge, we also leverage pre-trained word embeddings (complementary to topics) 
	accumulated in \texttt{WordPool} during lifelong learning. Essentially, we pool word embedding representation for every word $v_i$ learned while topic modeling over a stream of document collections from several domains. Thus, we have in total $T$ number of embeddings (encoding different semantics) for a word $v_i$ in \texttt{WordPool}, if the word appears in all the past collections. 
	Following \citeauthor{pankajgupta:2019iDocNADEe} \yrcite{pankajgupta:2019iDocNADEe}, we introduce prior knowledge in form of pre-trained word embeddings $[{\bf E}^{1},...,{\bf E}^{T}]$ in each hidden layer of DocNADE when analyzing  
	$\Omega^{T+1}$:
	\[{\bf h}({\bf v}_{<i}) = g({\bf c} + \sum_{q<i} {\bf W}_{:, v_q} + \sum_{q<i}\sum_{t=1}^{T} \lambda^{t}_{EmbTF} \ {\bf E}_{:, v_q}^{t})\]
	Observe that the topic learning for task $T+1$ is guided by an embedding vector ${\bf E}_{:, v_q}$ for the word $v_q$ from each of the $T$ domains (sources), where $\lambda^{t}_{EmbTF}$ is per-task transfer strength that controls the amount of prior (relevant) knowledge transferred to $T+1$ based on domain overlap with the past task $t$.   
	Discussed in section \ref{sec:docnade}, the word embedding representation ${\bf E}^{t} \in$ \texttt{WordPool} is obtained from the column-vectors of parameter ${\bf W}$ at the end of the task $t$.  
	
	\begin{algorithm}[t]
		\caption{\small Lifelong Learning Utilities}
		\label{algo:lifelongfunctions}
		\begin{center} 
			\begin{small} 
				\begin{algorithmic}[1]
					\FUNCTION {topic-learning ($\Omega, {\bf \Theta}$)}
					\STATE Build a DocNADE neural topic model: Initialize ${\bf \Theta}$
					\FOR {${\bf v} \in \Omega_{train}$}
					\STATE Forward-pass: 
					\STATE Compute loss, $\mathcal{L}({\bf v}  ; {\bf \Theta}) \gets$ compute-NLL(${\bf v}$, ${\bf \Theta}$)
					\STATE Backward-pass: 
					\STATE Minimize $\mathcal{L}({\bf v}  ; {\bf \Theta})$ using stochastic gradient-descent
					\ENDFOR
					\STATE Compute perplexity PPL of test set $\Omega_{test}$:
					\STATE $\mbox{PPL} \gets \exp\Large(\frac{1}{|\Omega_{test}|} \sum_{{\bf v} \in \Omega_{test}} \frac{\mathcal{L}({\bf v} ; {\bf \Theta})}{|{\bf v}|}\Large)$
					\STATE return PPL, ${\bf \Theta}$
					\ENDFUNCTION
					\item[]
					\FUNCTION{compute-NLL (${\bf v}, {\bf \Theta}$, \texttt{LNTM} = \{\})}
					\STATE Initialize ${\bf a} \gets {\bf c}$ and $ p({\bf v}) \gets 1$
					\FOR {word $i \in [1,..., N]$}
					\STATE  ${\bf h}_{i} ({\bf v}_{<i})  \gets g({\bf a}$), where $g$ = \{sigmoid, tanh\}
					\STATE $p(v_{i}=w | {\bf v}_{<i}) \gets \frac{\exp (b_w + {\bf U}_{w,:} {\bf h}_i ({\bf v}_{<i}))}{\sum_{w'} \exp (b_{w'} + {\bf U}_{w',:} {\bf h}_i ({\bf v}_{<i}))}$
					\STATE $ p({\bf v}) \gets  p({\bf v}) p(v_{i} | {\bf v}_{<i})$
					\STATE Compute pre-activation at $i^{th}$step: ${\bf a} \gets {\bf a} + {\bf W}_{:, v_{i}}$
					\IF  {\texttt{EmbTF} in  \texttt{LNTM}} 
					\STATE Get word-embedding vectors for $v_i$ from \texttt{WordPool}:  
					\STATE ${\bf a} \gets {\bf a} + \sum_{t=1}^{T} \lambda_{EmbTF}^t \ {\bf W}_{:, v_{i}}^{t}$	   
					\ENDIF
					\ENDFOR
					\STATE return $- \log p({\bf v} ; {\bf \Theta})$
					\ENDFUNCTION
					\item[]
					\FUNCTION {distill-documents (${\bf \Theta}^{T+1}$, $\mbox{PPL}^{T+1}$, [$\Omega^{1},..., \Omega^{T}$])}
					\STATE Initialize a set of selected documents: $\Omega_{aug}^{T+1} \gets \{\}$ 
					\FOR {task $t \in [1,...,T]$ and document ${\bf v}^t \in \Omega^{t}$}
					\STATE $\mathcal{L}({\bf v}^t ; {\bf \Theta}^{T+1}) \gets $ compute-NLL(${\bf v}^t$, ${\bf \Theta}^{T+1}$,  \texttt{LNTM} =\{\})
					\STATE $\mbox{PPL}({\bf v}^t ; {\bf \Theta}^{T+1}) \gets \exp(\frac{\mathcal{L}({\bf v}^t ; {\bf \Theta}^{T+1})}{|{\bf v}^{t}|})$
					\STATE Select document ${\bf v}^t$ for augmentation in task $T+1$:
					\IF {$\mbox{PPL}({\bf v}^t ; {\bf \Theta}^{T+1}) \le \mbox{PPL}^{T+1}$}
					\STATE Document selected: $\Omega_{aug}^{T+1} \gets  \Omega_{aug}^{T+1} \cup ({\bf v}^t, t)$
					\ENDIF
					\ENDFOR
					\STATE return $\Omega_{aug}^{T+1}$
					\ENDFUNCTION
				\end{algorithmic}
		\end{small} 
	\end{center}
\end{algorithm}
Figure \ref{fig:lifelongNTM} (component \circled{2}), Algorithm \ref{algo:lifelonglearning} (lines \#7-9) and Algorithm \ref{algo:lifelongfunctions} (lines \#20-23) illustrate the mechanism of topic modeling (DocNADE) with pre-trained word embeddings ${\bf E}$ from several sources (i.e., multi-source transfer learning) when learning topics ${\bf Z}^{T+1}$ for the future task $T+1$. 
\begin{figure*}[t]
	\begin{center}
		\includegraphics[scale=0.72]{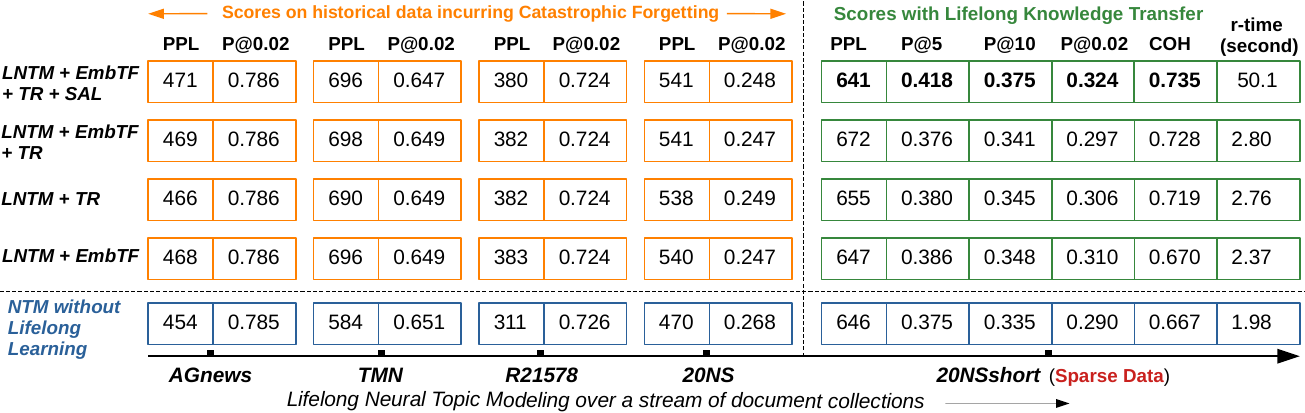}
		\vskip -0.1in
		\caption{PPL, P@R (precision@Recall), COH and r-time of LNTM system on future task, i.e., \texttt{20NSshort} over the stream {\bf S1}}
		\label{fig:20NSshortscores}
	\end{center}
\end{figure*}
\begin{figure*}[t]
	\begin{center}
		\includegraphics[scale=0.72]{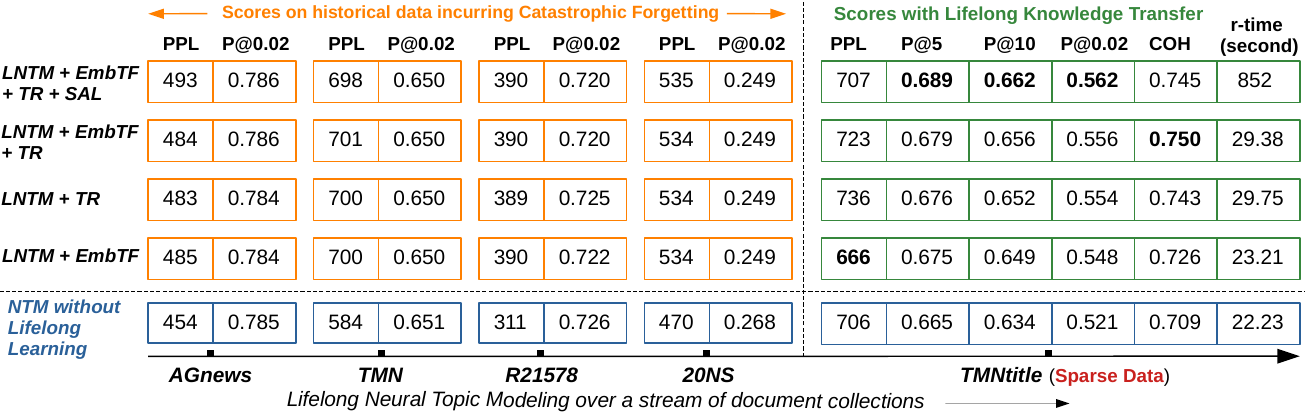}
		\vskip -0.1in
		\caption{PPL, P@R (precision@Recall), COH  and r-time of LNTM system on future task, i.e., \texttt{20TMNtitle} over the stream {\bf S2}}
		\label{fig:TMNtitlescores}
	\end{center}
	\vskip -0.2in
\end{figure*}
\begin{figure*}[t]
	\begin{center}
		\includegraphics[scale=0.72]{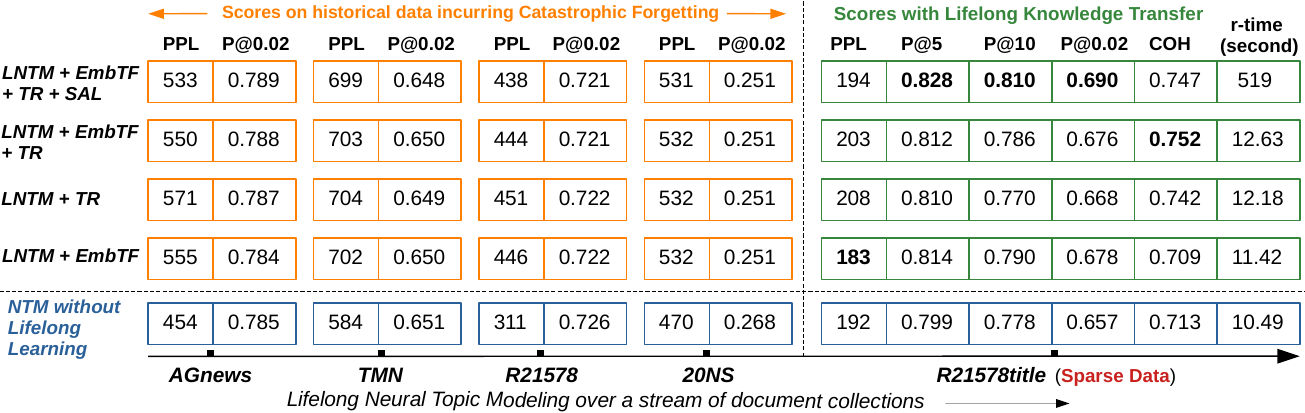}
		\vskip -0.1in
		\caption{PPL, P@R (precision@Recall), COH and r-time of LNTM system on future task, i.e., \texttt{R21578title} over the stream {\bf S3}}
		\label{fig:R21578titlescores}
	\end{center}
\end{figure*}

\begin{figure*}[t]
	\centering
	\begin{subfigure}{0.33\textwidth}
		\centering
		\begin{tikzpicture}[scale=0.60][baseline]
			\begin{axis}[
				x label style={at={(axis description cs:0.5,-0.05)},anchor=north},
				xlabel={\textbf{Fraction of Retrieved Documents (Recall)}},
				ylabel={\textbf{Precision (\%)}},
				xmin=-0.2, xmax=8.2,
				ymin=0.09, ymax=0.51,
				/pgfplots/ytick={.10,.14,...,.51},
				xtick={0,1,2,3,4,5,6,7,8},
				xticklabels={0.001, 0.002, 0.005, 0.01, 0.02, 0.05, 0.1, 0.2, 0.3},
				x tick label style={rotate=45,anchor=east},
				legend pos=north east,
				ymajorgrids=true, xmajorgrids=true,
				grid style=dashed,
				]
				\addplot[
				color=blue,
				mark=square,
				]
				plot coordinates {
					(0, 0.501)
					(1, 0.471)
					(2, 0.418)
					(3, 0.375)
					(4, 0.324)
					(5, 0.267)
					(6, 0.215)
					(7, 0.163)
					(8, 0.134)
				};
				\addlegendentry{\texttt{LNTM-all}}
				
				\addplot[
				color=orange,
				mark=triangle,
				]
				plot coordinates {
					(0, 0.470)
					(1, 0.418)
					(2, 0.375)
					(3, 0.335)
					(4, 0.290)
					(5, 0.242)
					(6, 0.199)
					(7, 0.153)
					(8, 0.127)    
				};
				\addlegendentry{\texttt{NTM}}
				
				\addplot[
				color=green,
				mark=*,
				]
				plot coordinates {
					(0, 0.388)
					(1, 0.378)
					(2, 0.327)
					(3, 0.281)
					(4, 0.236)
					(5, 0.188)
					(6, 0.154)
					(7, 0.123)
					(8, 0.107)
				};
				\addlegendentry{\texttt{EmbSUM}}
				
				\addplot[
				color=black,
				mark=square,
				]
				plot coordinates {
					(0, 0.381)
					(1, 0.377)
					(2, 0.343)
					(3, 0.309)
					(4, 0.272)
					(5, 0.219)
					(6, 0.178)
					(7, 0.138)
					(8, 0.116)
				};
				\addlegendentry{\texttt{zero-shot}}

				\addplot[
				color=cyan,
				mark=*,
				]
				plot coordinates {
					(0, 0.422)
					(1, 0.408)
					(2, 0.369)
					(3, 0.334)
					(4, 0.287)
					(5, 0.235)
					(6, 0.191)
					(7, 0.146)
					(8, 0.122)
				};
				\addlegendentry{\texttt{data-augment}}
			\end{axis}
		\end{tikzpicture}
		\caption{{\textbf{IR:}} 20NSshort} \label{fig:lifelongIR20NSshort}
	\end{subfigure}
	\begin{subfigure}{0.33\textwidth}
		\centering
		\begin{tikzpicture}[scale=0.60][baseline]
			\begin{axis}[
				x label style={at={(axis description cs:0.5,-0.05)},anchor=north},
				xlabel={\textbf{Fraction of Retrieved Documents  (Recall)}},
				ylabel={\textbf{Precision (\%)}},
				xmin=-0.2, xmax=8.2,
				ymin=0.24, ymax=0.67,
				/pgfplots/ytick={.25,.30,...,.7},
				xtick={0,1,2,3,4,5,6,7,8},
				xticklabels={0.001, 0.002, 0.005, 0.01, 0.02, 0.05, 0.1, 0.2, 0.3},
				x tick label style={rotate=48,anchor=east},
				legend pos=south west,
				ymajorgrids=true, xmajorgrids=true,
				grid style=dashed,
				]
				\addplot[
				color=blue,
				mark=square,
				]
				plot coordinates {
					(0, 0.649)
					(1, 0.633)
					(2, 0.609)
					(3, 0.588)
					(4, 0.562)
					(5, 0.515)
					(6, 0.464)
					(7, 0.391)
					(8, 0.336)
				};
				\addlegendentry{\texttt{LNTM-all}}
				
				\addplot[
				color=orange,
				mark=triangle,
				]
				plot coordinates {
					(0, 0.620)
					(1, 0.600)
					(2, 0.573)
					(3, 0.549)
					(4, 0.524)
					(5, 0.480)
					(6, 0.435)
					(7, 0.372)
					(8, 0.324)
				};
				\addlegendentry{\texttt{NTM}}
				
				\addplot[
				color=green,
				mark=*,
				]
				plot coordinates {
					(0, 0.633)
					(1, 0.610)
					(2, 0.574)
					(3, 0.544)
					(4, 0.512)
					(5, 0.461)
					(6, 0.411)
					(7, 0.343)
					(8, 0.296)
				};
				\addlegendentry{\texttt{EmbSUM}}
				
				\addplot[
				color=black,
				mark=square,
				]
				plot coordinates {
					(0, 0.608)
					(1, 0.581)
					(2, 0.547)
					(3, 0.517)
					(4, 0.487)
					(5, 0.444)
					(6, 0.405)
					(7, 0.352)
					(8, 0.312)
				};
				\addlegendentry{\texttt{zero-shot}}
				
				\addplot[
				color=cyan,
				mark=*,
				]
				plot coordinates {
					(0, 0.616)
					(1, 0.603)
					(2, 0.573)
					(3, 0.546)
					(4, 0.517)
					(5, 0.475)
					(6, 0.434)
					(7, 0.374)
					(8, 0.328)
				};
				\addlegendentry{\texttt{data-augment}}
			\end{axis}
		\end{tikzpicture}
		\caption{{\textbf{IR:}} TMNtitle} \label{fig:lifelongIRTMNtitle}
	\end{subfigure}
	\begin{subfigure}{0.33\textwidth}
		\centering
		\begin{tikzpicture}[scale=0.60][baseline]
			\begin{axis}[
				x label style={at={(axis description cs:0.5,-0.05)},anchor=north},
				xlabel={\textbf{Fraction of Retrieved Documents (Recall)}},
				ylabel={\textbf{Precision (\%)}},
				xmin=-.02, xmax=8.2,
				ymin=0.34, ymax=0.86,
				/pgfplots/ytick={.35,.45,...,.85},
				xtick={0,1,2,3,4,5,6,7,8},
				xticklabels={0.001, 0.002, 0.005, 0.01, 0.02, 0.05, 0.1, 0.2, 0.3},
				x tick label style={rotate=48,anchor=east},
				legend pos=south west,
				ymajorgrids=true, xmajorgrids=true,
				grid style=dashed,
				]
				\addplot[
				color=blue,
				mark=square,
				]
				plot coordinates {
					(0, 0.810)
					(1, 0.790)
					(2, 0.760)
					(3, 0.727)
					(4, 0.690)
					(5, 0.637)
					(6, 0.591)
					(7, 0.530)
					(8, 0.474)
				};
				\addlegendentry{\texttt{LNTM-all}}
				
				\addplot[
				color=orange,
				mark=triangle,
				]
				plot coordinates {
					(0, 0.778)
					(1, 0.759)
					(2, 0.728)
					(3, 0.695)
					(4, 0.657)
					(5, 0.600)
					(6, 0.551)
					(7, 0.494)
					(8, 0.436)
				};
				\addlegendentry{\texttt{NTM}}
				
				\addplot[
				color=green,
				mark=*,
				]
				plot coordinates {
					(0, 0.758)
					(1, 0.729)
					(2, 0.681)
					(3, 0.641)
					(4, 0.595)
					(5, 0.534)
					(6, 0.488)
					(7, 0.440)
					(8, 0.388)
				};
				\addlegendentry{\texttt{EmbSUM}}
				
				\addplot[
				color=black,
				mark=square,
				]
				plot coordinates {
					(0, 0.760)
					(1, 0.738)
					(2, 0.703)
					(3, 0.674)
					(4, 0.638)
					(5, 0.588)
					(6, 0.543)
					(7, 0.483)
					(8, 0.426)
				};
				\addlegendentry{\texttt{zero-shot}}
				
				\addplot[
				color=cyan,
				mark=*,
				]
				plot coordinates {
					(0, 0.762)
					(1, 0.740)
					(2, 0.710)
					(3, 0.680)
					(4, 0.645)
					(5, 0.593)
					(6, 0.544)
					(7, 0.487)
					(8, 0.428)
				};
				\addlegendentry{\texttt{data-augment}}
			\end{axis}
		\end{tikzpicture}
		\caption{{\textbf{IR:}} R21578title} \label{fig:lifelongIRR21578title}
	\end{subfigure}\hfill
	\caption{Precision-recall curve on 
		three future task datasets at different recall fractions. \texttt{LNTM-all} $\rightarrow$ \texttt{LNTM} + \texttt{EmbTF} + \texttt{TR} + \texttt{SAL}}
	\label{fig:lifelongIRplots}
\end{figure*}

{\bf \circled{3} Selective-Data Augmentation Learning} (\texttt{SAL}): 
Beyond the weight-based approaches in LML, the data-based approaches \cite{DBLP:journals/connection/Robins95} augment the training data of a future task with the data collected from the past tasks, allowing for (a) multi-task learning (MTL) \cite{DBLP:conf/icml/CollobertW08,DBLP:journals/corr/Ruder17a} to share representations among tasks and (b) minimizing catastrophic forgetting by data replay (augmentation).    
However, the data augmentation (DA) approaches are inefficient when the data collection grows large 
and often penalize positive transfer in MTL  due to domain shifts in the stream of data over lifetime.    

Our approach of SAL works in the following two steps:

{\bf Step 1} {\it Document Distillation} (Algorithm \ref{algo:lifelonglearning}: line \#17 and Algorithm \ref{algo:lifelongfunctions}: lines \#27-38):  Given document collections [$\Omega^{1}$,...,$\Omega^{T}$] of the past tasks, we ignore documents found not relevant in modeling a future task due to domain shifts. 
To do so, we first build a topic model with parameters ${\bf \Theta^{T+1}}$ over $\Omega^{T+1}$ and compute an average perplexity ($\mbox{PPL}^{T+1}$) score on its test set $\Omega^{T+1}_{test}$. 
Then, we prepare an augmented set $\Omega^{T+1}_{aug} \subset [\Omega^{1},...\Omega^{T}]$ such that each document ${\bf v}^t \in \Omega^{T+1}_{aug}$ of a past task $t$ satisfies: $\mbox{PPL}({\bf v}^{t};{\bf \Theta}^{T+1})$ $\le$ $\mbox{PPL}^{t+1}$.  In essence, this unsupervised document distillation scheme detects domain-overlap in the past and future tasks based on representation ability of ${\bf \Theta}^{T+1}$ for documents of the past. 

{\bf Step 2} {\it Selective Co-training} (Algorithm \ref{algo:lifelonglearning}: lines \#18-20): We re-train topic modeling over $\Omega^{T+1}$ simultaneously using  $\Omega^{T+1}_{aug}$, leveraging topical homologies in (selective) documents of the past and future tasks, as:
\begin{align*}
	\begin{split}
		\Delta_{SAL}  =&    \sum_{({\bf v}^{t}, t) \in \Omega^{T+1}_{aug}} \lambda_{SAL}^t \ \mathcal{L}( {\bf v}^{t} ;{\bf \Theta}^{T+1})\\
		\mathcal{L}(\Omega^{T+1} ; {\bf \Theta}^{T+1})  = & \sum_{{\bf v} \in \Omega^{T+1}} \mathcal{L}( {\bf v} ;{\bf \Theta}^{T+1})  +  \ \Delta_{SAL}
	\end{split}
\end{align*}
Here, $\lambda_{SAL}^t$ is per-task contribution that modulates influence of shared representations while co-training with selected documents of the past task $t$. 
The SAL approach jointly helps in transferring prior knowledge from several domains, minimizing catastrophic forgetting and reduce training time due to selective data replay over lifetime. 

{\bf Overall loss in LNTM framework:} Combining the different approaches within the proposed lifelong learning paradigm, 
the overall loss in modeling documents $\Omega^{T+1}$ being the future (new) task $T+1$ is given by:
\begin{equation*}
	\mathcal{L}(\Omega^{T+1} ; {\bf \Theta}^{T+1}) = \sum_{{\bf v} \in \Omega^{T+1}} \mathcal{L}( {\bf v} ;{\bf \Theta}^{T+1})  +  \  \Delta_{TR} +  \ \Delta_{SAL}
\end{equation*}
{\it Computation complexity of LNTM}: 
In DocNADE (section \ref{sec:docnade}) without LNTM, the complexity of computing the joint distribution $p({\bf v})$ is in $O(DH + KDH)$. The complexity of computing $\Delta_{TR}$ and $\Delta_{SAL}$  are in $O(KH+KH)$ and $O(DH + KDH)$, respectively. The overall complexity of \texttt{LNTM} = \{\texttt{EmbTF}, \texttt{TR}, \texttt{SAL}\}  is in 
$O(DH + KDH + KH+KH + DH + KDH) \sim O(DH + KDH + KH)$.  

{\bf Phase 2: Lifelong Knowledge Accumulation:} For each topic modeling task $t$, the phase 1 generates knowledge in form of topic and word embeddings that is respectively accumulated in \texttt{TopicPool} $\gets \mbox{row-vectors} ({\bf W} \in {\bf \Theta}^{t}$) and  \texttt{WordPool} $\gets \mbox{column-vectors}({\bf W} \in {\bf \Theta}^{t}$). Additionally, each  decoding parameter ${\bf U} \in {\bf \Theta}^{t}$ is retained to be used in minimizing catastrophic forgetting 
(i.e., $\Delta_{TR}$).   

\section{Experiments and Analysis}\label{sec:experiments}
{\bf Streams of Document Collections:} 
To demonstrate the applicability of our proposed LNTM framework, we prepare a stream of document collections consisting of 
four long-text (high-resource) corpora in sequence:  \texttt{AGnews}, \texttt{TMN}, \texttt{R21578} and \texttt{20NS} (20NewsGroups), and three short-text (low-resource) corpora $\Omega^{T+1}$ as {\bf future tasks} $T+1$:  
\texttt{20NSshort},  \texttt{TMNtitle} and \texttt{R21578title}. 
Thus, we perform lifelong topic learning over following three streams:\\ 
{\bf S1}:  {\small \texttt{AGnews} $\rightarrow$ \texttt{TMN} $\rightarrow$ \texttt{R21578} $\rightarrow$ \texttt{20NS} $\rightarrow$ \texttt{20NSshort}}\\
{\bf S2}:  {\small \texttt{AGnews} $\rightarrow$ \texttt{TMN} $\rightarrow$ \texttt{R21578} $\rightarrow$ \texttt{20NS} $\rightarrow$ \texttt{TMNtitle}}\\
{\bf S3}:  {\small \texttt{AGnews} $\rightarrow$ \texttt{TMN} $\rightarrow$ \texttt{R21578} $\rightarrow$ \texttt{20NS} $\rightarrow$ \texttt{R21578title}}\\
such that we demonstrate improved topic modeling for the three sparse document collections ($\Omega^{T+1}$) at $T+1$. 
The order of $\Omega$s is based on their decreasing sizes. 
See the {\it supplementary} for data description and domain overlap.

{\bf Baselines:} Discussed in section \ref{sec:docnade}, we adopt DocNADE (NTM: a neural topic modeling tool) and compare it with the proposed framework \texttt{LNTM} = \{\texttt{EmbTF}, \texttt{TR}, \texttt{SAL}\}. 
Moreover, we show topic learning in {\it zero-shot}, {\it few-shot} and {\it data augmentation} settings in the following section. 

{\bf Reproducibility:} 
PPL (Algorithm \ref{algo:lifelongfunctions}: line \#10) is used for model selection and adjusting parameters ${\bf \Theta}^{t}$ and hyper-parameters ${\bf \Phi}^{t}$. 
See the {\it supplementary} for the hyper-parameters settings. 
Figures \ref{fig:20NSshortscores}, \ref{fig:TMNtitlescores} and \ref{fig:R21578titlescores} show average run-time (r-time) for each training epoch of different LNTM approaches, run on an  NVIDIA Tesla K80 Processor (RAM: 12 GB) to a maximum of 100 epochs.  

To evaluate the capabilities of LNTM framework, we employ three measures: precision@recall (P@R) in information retrieval (IR) task for {\it document representation}, topic coherence (COH) for {\it topic quality}  and perplexity (PPL) for {\it generative performance} of topic modeling over lifetime.  

\subsection{Document Representation via Retrieval (IR)}\label{sec:IR}
To evaluate the quality of document representation learned within LNTM, 
we perform an unsupervised document retrieval task for each collection over lifetime. 
In doing so, we compute average P@R on the test set for a task $t$, 
where each test document is treated as a test query to retrieve a fraction/top $R$ of the closest documents in the training set. 
We compute cosine similarity between document vectors (i.e., the last hidden ${\bf h}_{D}$ of DocNADE) and average 
the number of retrieved documents with the same label as the query.
Figures \ref{fig:20NSshortscores}, \ref{fig:TMNtitlescores} and \ref{fig:R21578titlescores} show P@5, P@10 and P@0.02 on the all test collections of the streams {\small \bf S1}, {\small \bf S2} and {\small \bf S3}, respectively accounting for knowledge transfer and forgetting. 

{\it Precision@Recall on future tasks}:
Figures \ref{fig:20NSshortscores}, \ref{fig:TMNtitlescores} and \ref{fig:R21578titlescores} report P@5, P@10 and P@0.02 scores (green boxes) on three future tasks:  
{\small \texttt{20NSshort}}, {\small \texttt{TMNtitle}} and {\small \texttt{R21578title}}, respectively leveraging prior knowledge over lifetime.  
Compared to NTM without lifelong learning (blue boxes), all the proposed approaches: \texttt{EmbTF}, \texttt{TR} and \texttt{SAL} (green boxes) within LNTM
outperform it for all the future tasks, e.g., 
P@0.02: ($.324$ vs $.290$), ($.562$ vs $.521$) and ($.690$ vs $.657$) on {\small \texttt{20NSshort}}, {\small \texttt{TMNtitle}} and {\small \texttt{R21578title}}, respectively due to {\small \texttt{LNTM}+\texttt{EmbTF}+\texttt{TR}+\texttt{SAL}}.  Observe that the \texttt{SAL} leads to higher gains when combined with the other LNTM approaches, 
suggesting a positive knowledge transfer from both the past learning and document collections.   

{\it Precision@Recall on past tasks incurring forgetting} (orange boxes): 
To demonstrate the ability of LNTM framework in minimizing catastrophic forgetting, we also report P@R scores on 
the past tasks\footnote{Due to partially overlapping vocabulary in $\Omega$s over a stream, we overwrite column-vectors of ${\bf W}$$\in$${\bf \Theta}^{t}$ by column-vectors of ${\bf W}$$\in$${\bf \Theta}^{T+1}$ for all words $v_i$ appearing in both tasks $t$ and $T+1$}    
using parameters of a future task ${\bf \Theta}^{T+1}$. 
Figures \ref{fig:20NSshortscores}, \ref{fig:TMNtitlescores} and \ref{fig:R21578titlescores} report P@0.02 for each of the past tasks over lifetime using {\bf S1}, {\bf S2} and {\bf S3} streams, suggesting that the proposed approaches in LNTM help in preventing catastrophic forgetting. For each stream, compare scores in the orange and blue boxes column-wise correspondingly for each task.  
E.g., P@0.02 for {\small \texttt{TMN}} in {\bf S1}, {\bf S2} and {\bf S3} incurring forgetting are ($.647$ vs $.651$), ($.650$ vs $.651$) and ($.648$ vs $.651$), respectively advocating for representation capabilities of the future tasks for the past learning within LNTM.

{\bf Zero-shot and Data-augmentation Investigations:}
Additionally, we analyze representation capabilities of LNTM in zero-shot and data-augmentation settings, where 
we compute P@R on all future tasks $T+1$ respectively using parameters:  
(a) ${\bf \Theta}^{T}$ learned from the past task $T$ and no $\Omega^{T+1}$ used, and 
(b) ${\bf \Theta}^{T+1}$ learned on a future task by combining all document collections [$\Omega^{1},...,\Omega^{T+1}$] in a stream.  
Figures \ref{fig:lifelongIR20NSshort}, \ref{fig:lifelongIRTMNtitle} and \ref{fig:lifelongIRR21578title} show precision-recall plots for {\small \texttt{20NSshort}}, {\small \texttt{TMNtitle}} and {\small \texttt{R21578title}} datasets, respectively.   
Observe that the proposed approach \texttt{LNTM-all} (i.e., {\small \texttt{LNTM} + \texttt{EmbTF} + \texttt{TR}+ \texttt{SAL}}) outperforms NTM (i.e., DocNADE without lifelong learning), zero-shot, data-augment and {\small \texttt{EmbSUM}} baselines at all the retrieval fractions. Here, {\small \texttt{EmbSUM}} represents a document by summing the embedding vectors of its words using Glove embeddings \cite{D14-1162}. 

\subsection{Topic Quality via Coherence (COH)}\label{sec:COH}
Beyond document representation, topic models essentially enable interpretability by generating topics (sets of key terms) that explain thematic structures hidden in document collections. Topics are often incoherent when captured in data sparsity (low-resource) settings, leading to restrict the interpretability. 
Thus, we compute topic coherence (COH) scores proposed by \citeauthor{DBLP:conf/wsdm/RoderBH15} \yrcite{DBLP:conf/wsdm/RoderBH15} to estimate the quality (meaningfulness of words) of topics captured within LNTM framework. 
Following \citeauthor{pankajgupta:2019iDocNADEe} \yrcite{pankajgupta:2019iDocNADEe}, we compute COH (Figures 
\ref{fig:20NSshortscores}, \ref{fig:TMNtitlescores} and \ref{fig:R21578titlescores}) on the top-10 words in each topic. The higher scores imply topic coherency.

{\it COH scores on future tasks}: Within LNTM, we show a gain of 10.2\% ($0.735$ vs $0.667$), 5.8\%($0.750$ vs $0.709$) and 5.5\%($0.752$ vs $0.713$) respectively on the three sparse datasets, suggesting quality topics discovered.

\begin{figure}[hpt]
	\begin{center}
		\includegraphics[scale=0.68]{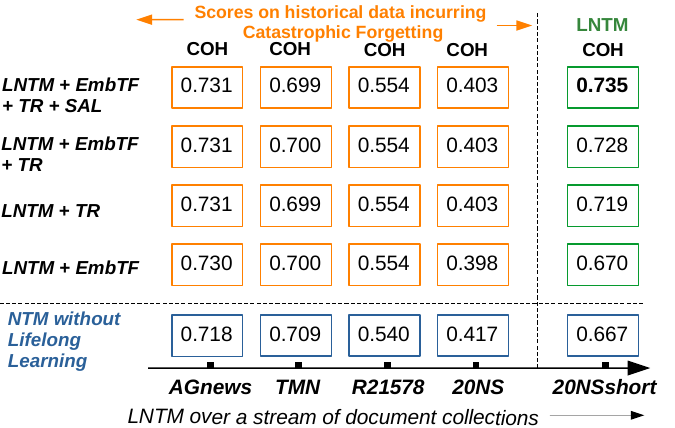}
		\vskip -0.1in
		\caption{COH  on future (\texttt{20NSshort})  and past tasks for {\bf S1}}
		\label{figsuppl:20NSshortscores}
	\end{center}
	\vskip -0.2in
\end{figure}

\begin{figure}[thp]
	\begin{center}
		\includegraphics[scale=0.68]{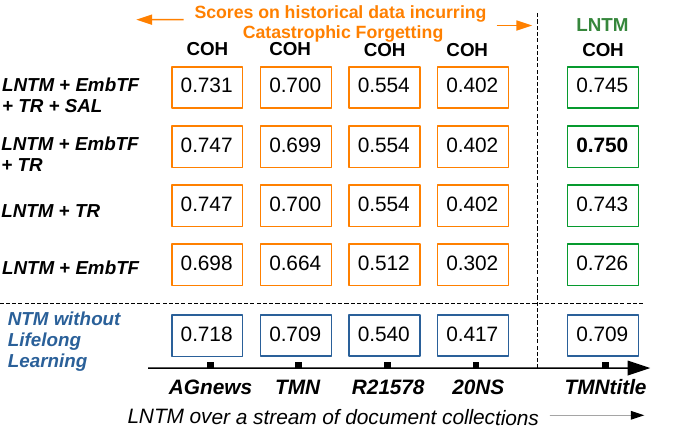}
		\vskip -0.1in
		\caption{COH  on future (\texttt{TMNtitle})  and past tasks for {\bf S2}}
		\label{figsuppl:TMNtitlescores}
	\end{center}
	\vskip -0.2in
\end{figure}

\begin{figure}[hpt]
	\begin{center}
		\includegraphics[scale=0.68]{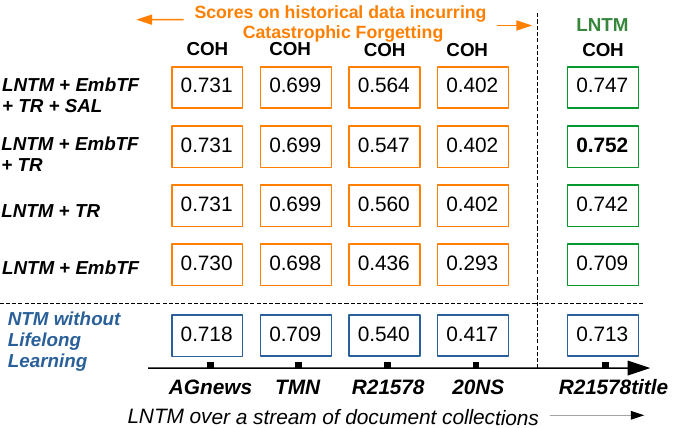}
		\vskip -0.1in
		\caption{COH  on future (\texttt{R21578title})  and past tasks for {\bf S3}}
		\label{figsuppl:R21578titlescores}
	\end{center}
	\vskip -0.2in
\end{figure}

Figures \ref{figsuppl:20NSshortscores}, \ref{figsuppl:TMNtitlescores} and \ref{figsuppl:R21578titlescores} show topic coherence (COH) scores on document collections in streams S1,S2 and S3, respectively.  We also show scores incurring forgetting on past tasks in each of the three streams.  Our proposed lifelong topic modeling framework reports gains in topic coherence scores for each of the target (future) tasks and also minimizes catastrophic forgetting on the past tasks.

\subsection{Generalization via Perplexity (PPL)}\label{sec:PPL}
To evaluate generative performance of topic models, 
we estimate the log-probabilities for unseen test documents $\Omega^{T+1}_{test}$ of the future tasks,
and compute the average held-out perplexity per word (Algorithm \ref{algo:lifelongfunctions}: line \#10). 
Note that lower the PPL (negative log-likelihood), better the topic model. 
Figures \ref{fig:20NSshortscores}, \ref{fig:TMNtitlescores} and \ref{fig:R21578titlescores} show PPL scores on all (test) document collections in the streams {\bf S1}, {\bf S2} and {\bf S3}, respectively.

{\it PPL on future tasks}: Figure \ref{fig:20NSshortscores} shows PPL scores on the future task using {\small \texttt{20NSshort}} without (blue boxes) and with (green boxes) lifelong settings. 
Compared to NTM, the configuration {\small \texttt{LNTM}+\texttt{EmbTF}+\texttt{TR}+\texttt{SAL}} reports an improved score of ($641$ vs $646$).  Similarly,  Figures \ref{fig:TMNtitlescores} and \ref{fig:R21578titlescores} depict that the generalization capability is boosted, i.e., ($666$ vs $706$) and ($183$ vs $192$) on {\small \texttt{TMNtitle}} and {\small \texttt{R21578title}}, respectively due to word-embedding based multi-domain multi-source knowledge transfer ({\small \texttt{LNTM}+\texttt{EmbTF}}) over lifetime. 

{\it PPL on past tasks incurring forgetting}(orange boxes): We also report PPL on the all  past document collections of the streams {\bf S1}, {\bf S2} and {\bf S3} using parameters ${\bf \Theta}^{T+1}$ of a future task. Comparing the proposed approaches of LNTM,  we observe that they retain PPL over lifetime for each document collection in each of the streams; however at the cost of forgetting 
due to sensitivity of the log-likelihood computation towards neural network parameters. Note that ${\bf \Theta}^{T+1}$ retains representation ability for all $t<T+1$ quantified by IR.

\begin{figure}[t]
	\centering
	\begin{center}
		\includegraphics[scale=0.58]{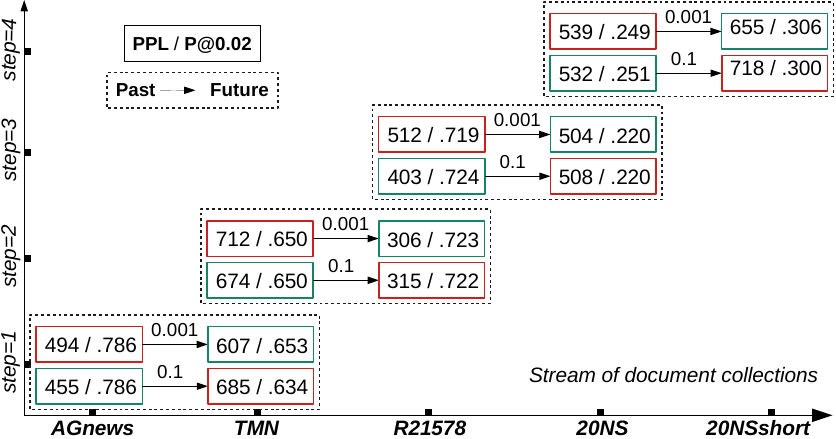}
		\caption{Ablation for $\lambda_{TR}$: Maximum knowledge transfer vs minimum catastrophic forgetting over lifetime using stream {\bf S1} }
		\label{fig:LNTManalysisforgettingvstransfer}
	\end{center}
\end{figure}

\begin{figure}[t]
	\centering
	\center
	\includegraphics[scale=0.67]{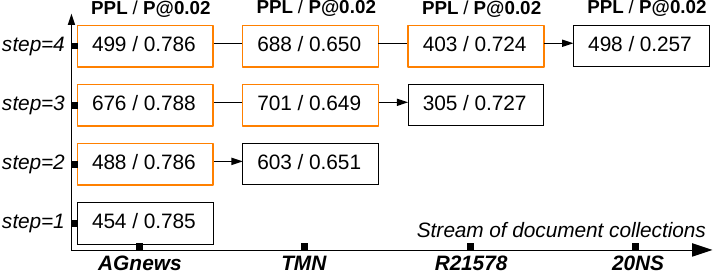}	
	\caption{Illustration of LNTM (\texttt{LNTM-all}) for each task over stream {\bf S1}, where orange colored boxes indicate scores incurring forgetting while modeling a future task (gray boxes) at each step}
	\label{figsuppl:LNTManalysis}
\end{figure}

\subsection{Analysis: Quantitative and Qualitative}\label{sec:analysis}
{\bf Knowledge Transfer vs Forgetting:}
While learning within lifelong framework, there is a trade-off in knowledge transfer for future and forgetting of past. 
In Figure \ref{fig:LNTManalysisforgettingvstransfer}, we provide an ablation over $\lambda_{TR} \in \{0.001, 0.1\}$ for \texttt{LNTM} = \{TR\} approach to show how $\lambda_{TR}$ regulates the trade-off.  Observe that the lower values of $\lambda_{TR}$ leads to maximizing knowledge transfer (green boxes) for a future task (within a gray box), however at the cost of forgetting (red) of past learning and vice-versa when $\lambda_{TR}$ increases. Here at each step (y-axis), we perform the ablation in pairs of document collections over the stream {\bf S1} and show PPL and P@R. The study suggests to set $\lambda_{TR}$ such that the trade-off is balanced. 

{\bf Lifelong Topic Learning over the stream {\bf S1}:}
Similar to Figures (\ref{fig:20NSshortscores}, \ref{fig:TMNtitlescores} and \ref{fig:R21578titlescores}), we additionally provide an illustration 
of lifelong topic learning over {\bf S1}, where each task in sequence is treated as a future task accounting for the trade-off and forgetting over lifetime. 
Figure \ref{figsuppl:LNTManalysis} provides illustration of LNTM (\texttt{LNTM-all}) for each dataset (as future task in gray box) in the streams of document collection used. 
Here, we show scores: PPL and P@0.02 of generalization and IR task, where the y-axis indicates each step of the lifelong learning process of topic modeling over a stream of document collections. Observe that the orange color box indicates scores incurring forgetting while modeling a target (in gray box).   Once the step 4 is executed, we use three sparse targets (as step 5) to show applicability of lifelong topic modeling to address data-sparsity issues. 

{\bf Qualitative Topics:}
Table \ref{tab:topiccoherencescores} shows topics (top-5 words) captured on \texttt{TMNtitle} (sparse) document collection of the stream {\bf S2}, extracted using row-vectors of ${\bf W} \in {\bf \Theta}^{T+1}$.  Observe that NTM generates incoherent topics (terms marked in red); however the two topics (T1 and T2) becomes coherent within LNTM framework, representing thematic structures about  \textit{product-line} and \textit{disaster}, respectively. It suggest that the quality of topics is improved due to a positive transfer of knowledge via \texttt{EmbTF}, \texttt{TR} and \texttt{SAL} approaches. 
\begin{table}[t]
	\vskip -0.1in
	\caption{Analysis: Qualitative topics of \texttt{TMNtitle}}
	\label{tab:topiccoherencescores}
	\vskip -0.1in
	\begin{center}
		\renewcommand*{\arraystretch}{1.2}
		\resizebox{.485\textwidth}{!}{
			\begin{tabular}{c|l}
				\toprule
				\multicolumn{1}{c|}{\bf Model}      & \multicolumn{1}{c}{\bf Topic-words (Top 5)}     \\ 
				\midrule
				\multirow{2}{*}{\texttt{NTM}}			&  T1:  {\color{red}nuclear}, {\color{red}break}, jobs, {\color{red}afghanistan}, ipad		\\ 
				&	T2:  {\color{red}gulf}, {\color{red}bruins}, japanese, {\color{red}michigan}, radiation \\ \hline

				\multirow{2}{*}{\texttt{LNTM} + \texttt{TR}}	&   T1:  {\color{red}arts}, android, iphone, tablet, ipad	\\  
				&   T2:  {\color{red}rail}, medicare, wildfire, radioactive, {\color{red}recession} \\  \cdashline{1-2}  
				
				\multirow{2}{*}{\texttt{LNTM-all}}			& T1:  linkedin, android, tablet, ipad, iphone	\\  
				&  T2:  tornadoes, fukushima, radioactive, radiation, medicare \\        
				\bottomrule
		\end{tabular}}
	\end{center}
	\vskip -0.2in
\end{table}

\section{Conclusion}
We have presented a novel lifelong neural topic modeling framework that models a stream of document collections and exploits prior knowledge from several domains over lifetime in form of pre-trained topics, word embeddings and generative homologies in historical collections. Experimental results show that our proposed approaches of joint topic regularization, selective-data augmented learning and word-embedding guided topic learning within the lifelong framework help modeling three sparse datasets, quantified by information retrieval, topic coherence and generalization.     

\section*{Acknowledgment}

This research was supported by Bundeswirtschaftsministerium (bmwi.de), grant
01MD19003E (PLASS-Platform for Analytical Supply Chain Management Services (plass.io)) at Siemens AG- CT Machine Intelligence, Munich Germany.

\bibliography{icml2020}
\bibliographystyle{icml2020}

\appendix

\section{Data Description}\label{secsuppl:datadescription}

\begin{table}
	\center
	\caption{Data statistics: Document collections used in lifelong topic modeling.  Symbols-  $K$: vocabulary size,  $L$: average text length (\#words), $C$: number of classes and  $k$: thousand. For short-text,  $L$$<$$15$. We use $\mathcal{T}^1$, $\mathcal{T}^2$ and $\mathcal{T}^3$ are treated as target corpora for future tasks $T+1$ of topic modeling and $\mathcal{S}^1$-$\mathcal{S}^4$ are used as historical  corpora in the stream of document collections.}
	\label{tabsuppl:datadescription}
	\vskip 0.1in
	\renewcommand*{\arraystretch}{1.1}
	\resizebox{.49\textwidth}{!}{
		\begin{tabular}{r|r|rrrrrr}
			\toprule
			\multicolumn{1}{c|}{\bf ID} & \multicolumn{1}{c|}{\bf Data} &  \multicolumn{1}{c}{\bf Train} &  \multicolumn{1}{c}{\bf Val} & \multicolumn{1}{c}{\bf Test} &  \multicolumn{1}{c}{$K$}   & \multicolumn{1}{c}{\bf L} & \multicolumn{1}{c}{\bf C}   \\ \midrule 
			$\mathcal{T}^1$	&    20NSshort             & 1.3k       & 0.1k        & 0.5k           &    1.4k           &  13.5        &   20                   \\
			$\mathcal{T}^2$	&    TMNtitle                & 22.8k    &  2.0k     & 7.8k              &       2k           &   4.9          &     7     \\
			$\mathcal{T}^3$	&    R21578title            &  7.3k     &  0.5k     & 3.0k            &   2k                  &   7.3          &     90   \\
			$\mathcal{S}^1$  	& AGNews              & 118k & 2.0k &   7.6k &        5k         &  38       &     4         \\
			$\mathcal{S}^2$  	&   TMN                 & 22.8k  &  2.0k & 7.8k  &      2k         &    19          &       7     \\
			$\mathcal{S}^3$  	&  R21578              &  7.3k&  0.5k & 3.0k  &      2k               &   128       &   90       \\
			$\mathcal{S}^4$  	& 20NS                  & 7.9k & 1.6k & 5.2k  &     2k          &   107.5     &   20        \\
			\bottomrule          
	\end{tabular}}
\end{table}

\begin{table}[t]
	\centering
	\caption{Illustration of Domain-overlap in pairs of corpora, when used in source-target settings. 
		$\mathcal{I}$: Identical,  $\mathcal{R}$: Related and $\mathcal{D}$: Distant domains determined based on overlap in labels}
	\label{tabsuppl:domainoverlap}
	\vskip 0.1in
	\renewcommand*{\arraystretch}{1.1}
	\resizebox{.2\textwidth}{!}{
		\begin{tabular}{|c||c|c|c|} 
			\toprule
			& $\mathcal{T}^1$	& $\mathcal{T}^2$ 	& $\mathcal{T}^3$  \\ \hline \hline
			$\mathcal{S}^1$	& $\mathcal{R}$ 			& $\mathcal{R}$  			& $\mathcal{D}$ 			  \\ \hline
			$\mathcal{S}^2$	& $\mathcal{R}$  			& $\mathcal{I}$ 			& $\mathcal{D}$  			\\ \hline
			$\mathcal{S}^3$	& $\mathcal{D}$  			& $\mathcal{D}$ 			& $\mathcal{I}$				 \\ \hline
			$\mathcal{S}^4$	& $\mathcal{I}$ 			& $\mathcal{R}$ 			& $\mathcal{D}$ 		 	  \\ \hline
	\end{tabular}}
\end{table}

Discussed in section 3, we perform lifelong topic learning over following three streams:\\ 
{\small 
	{\bf S1}:  \texttt{AGnews} $\rightarrow$  \texttt{TMN} $\rightarrow$  \texttt{R21578} $\rightarrow$  \texttt{20NS} $\rightarrow$  \texttt{20NSshort}\\  
	{\bf S2}:  \texttt{AGnews} $\rightarrow$  \texttt{TMN} $\rightarrow$  \texttt{R21578} $\rightarrow$  \texttt{20NS} $\rightarrow$  \texttt{TMNtitle}\\  
	{\bf S3}:  \texttt{AGnews} $\rightarrow$  \texttt{TMN} $\rightarrow$  \texttt{R21578} $\rightarrow$  \texttt{20NS} $\rightarrow$  \texttt{R21578title}  
}

Each stream of document collections consisting of four long-text (high-resource) corpora in sequence: \texttt{AGnews}, 
\texttt{TMN}, \texttt{R21578} and \texttt{20NS} (20NewsGroups), and three short-text (low-resource, sparse) corpora $T+1$ as future (target) tasks $T + 1$: \texttt{20NSshort}, \texttt{TMNtitle} and  \texttt{R21578title}. 

Following is the description of document collections used in this work: 
\begin{enumerate}
	\item \texttt{20NSshort}: We take documents from 20NewsGroups data, with document size (number of words) less than 20.
	\item \texttt{TMNtitle}: Titles of the Tag My News (TMN) news dataset. 
	\item \texttt{R21578title}:  Reuters corpus, a collection of new stories from \url{nltk.corpus}. We take titles of the documents.
	\item \texttt{TMN}: The Tag My News (TMN) news dataset. 
	\item  \texttt{R21578}: Reuters corpus, a collection of new stories from \url{nltk.corpus}.   
	\item \texttt{AGnews}:  AGnews data sellection. 
	\item \texttt{20NS}: 20NewsGroups corpus, a collection of  news stories from \url{nltk.corpus}.  
\end{enumerate}

See Table \ref{tabsuppl:datadescription} for the description of each of the document collections used in our experiments. Observe that we employ sparse document collections as target datasets.  

Table \ref{tabsuppl:domainoverlap} suggests a domain overlap (in terms of labels) among the document collections used in transfer learning within neural topic modeling framework. The notations such as $\texttt{I}$, $\texttt{R}$ and $\texttt{D}$ represent domain overlap, where $\texttt{I}$ (identical): identical-domain in terms of labels in pair of datasets, $\texttt{R}$ (related):  related-domain due to partial overlap in labels, and $\texttt{D}$: distant-domain due to no overlap in labels of pair of document collections. 

See Table  \ref{tabsuppl:labelspace} for the label information for each of the document collections used in streams of information to model. 

To reproduce the scores reported, we have also provided the {\bf code} of the LNTM framework and {\bf pre-processed datasets}  used in our experiments. 

\section{Reproducibility: Hyper-parameter Settings}

In the following sections, we provide the hyper-parameter settings (search space) 
used to build topic models based on development set.

\subsection{Hyper-parameter settings for Generalization}
Table \ref{suppl:HyperparametersinPPL} provides hyper-parameters search space used within lifelong topic modeling framework for generalization task over lifetime. The models built are used further in extracting topics and computing topic coherence. 

\begin{table}[h]
	\centering
	\caption{Hyper-parameters search space in the Generalization task}
	\label{suppl:HyperparametersinPPL}
	\vskip 0.1in
	\resizebox{.4\textwidth}{!}{
		\begin{tabular}{c|c}
			\hline 
			{\bf Hyperparameter}               & {\bf Search Space} \\ \hline
			retrieval fraction        &    [0.02]                        \\
			learning rate        &    [{0.001}]                       \\
			hidden units (\#topics), $H$         &      [50, 200]               \\ 
			activation function ($g$)        &      {sigmoid}     \\
			iterations        &      [100]      \\
			$\lambda_{TR}$        &      [0.1, 0.01, 0.001]  \\ 
			$\lambda_{EmbTF}$        &      [1.0, 0.5, 0.1]  \\
			$\lambda_{SAL}$        &      [1.0, 0.5, 0.1]  \\   \bottomrule 
	\end{tabular}}
\end{table}

\subsection{Hyper-parameter settings for IR Task}

Table \ref{suppl:HyperparametersinIR} provides hyper-parameters search space used within lifelong topic modeling framework for information retrieval task over lifetime. 

\begin{table}[h]
	\centering
	\caption{Hyper-parameters search space in the IR task}
	\label{suppl:HyperparametersinIR}
	\vskip 0.1in
	\resizebox{.4\textwidth}{!}{
		\begin{tabular}{c|c}
			\hline 
			{\bf Hyperparameter}               & {\bf Search Space} \\ \hline
			retrieval fraction        &    [0.02]                        \\
			learning rate        &    [{0.001}]                       \\
			hidden units (\#topics), $H$         &      [50, 200]               \\ 
			activation function ($g$)        &      {tanh}     \\
			iterations        &      [100]      \\
			$\lambda_{TR}$        &      [0.1, 0.01, 0.001]  \\ 
			$\lambda_{EmbTF}$        &      [1.0, 0.5, 0.1]  \\
			$\lambda_{SAL}$        &      [1.0, 0.5, 0.1]  \\   \bottomrule 
	\end{tabular}}
\end{table}  

 \subsection{Optimal Configurations of $\lambda^{TR}$, $\lambda^{EmbTF}$, $\lambda^{SAL}$}

Tables \ref{optimalhyperparams:PPLtask} and \ref{optimalhyperparams:IRtask} provide the optimal (best) hyper-parameter setting for generalization and IR task, respectively for each of the three target datasets.  The hyper-parameters corresponds to the scores reported in the paper content. 

To reproduce the scores reported, we have also provided the {\bf code} of the LNTM framework and {\bf pre-processed datasets}  used in our experiments.

\begin{table*}[h]
	\caption{Generalization Task: Optimal settings of hyper-parameters ($\lambda_{TR}$ / $\lambda_{EmbTF}$ / $\lambda_{SAL}$) for each of the three streams where the datasets: \texttt{20NSshort}, \texttt{TMNtitle} and \texttt{R21578title} are treated as targets, respectively in each of the streams. The optimal hyper-parameters are obtained in joint training of three approaches: \texttt{TR}, \texttt{EmbTF} and \texttt{SAL} with the proposed LNTM framework.}
	\label{optimalhyperparams:PPLtask}
	\begin{center}
		\renewcommand*{\arraystretch}{1.2}
		\resizebox{.95\textwidth}{!}{
			\begin{tabular}{c|c|c|c|c|c}
				\toprule
				\multirow{2}{*}{\bf Target} & \multirow{2}{*}{\bf Stream} & \multicolumn{4}{c}{\bf Hyper-parameters ($\lambda_{TR}$ / $\lambda_{EmbTF}$ / $\lambda_{SAL}$) for Streams of Document Collections} \\ 
				& 			& \texttt{AGnews} 			  & \texttt{TMN} 		     & \texttt{R21578} 			& \texttt{20NS} \\ \midrule
				\texttt{20NSshort}     & {\bf S1}		&     0.001 / 0.1 / 1.0  		  & 	0.001 / 0.1 / 1.0	     & 		0.001 / 0.1 / 1.0	& 0.001 / 1.0 / 1.0 \\ \hline
				\texttt{TMNtitle}        & {\bf S2}		&     0.001  / 0.1 / 0.1 		  & 	0.001 / 1.0 / 1.0	     & 		0.001 / 0.1 / 0.1	& 0.001 / 0.1 / 0.1 \\ \hline
				\texttt{R21578title}    & {\bf S3}	&     0.001 / 0.1 / 0.1  		  & 	0.001 / 0.1 / 0.1	     & 		0.001 / 1.0 / 0.1	& 0.1 / 0.1 / 0.1 \\ \bottomrule
		\end{tabular}}
	\end{center}
\end{table*}

\begin{table*}[h]
	\caption{IR Task: Optimal settings of hyper-parameters ($\lambda_{TR}$ / $\lambda_{EmbTF}$ / $\lambda_{SAL}$) for each of the three streams where the datasets: \texttt{20NSshort}, \texttt{TMNtitle} and \texttt{R21578title} are treated as targets, respectively in each of the streams. The optimal hyper-parameters are obtained in joint training of three approaches: \texttt{TR}, \texttt{EmbTF} and \texttt{SAL} with the proposed LNTM framework.}
	\label{optimalhyperparams:IRtask}
	\begin{center}
		\renewcommand*{\arraystretch}{1.2}
		\resizebox{.95\textwidth}{!}{
			\begin{tabular}{c|c|c|c|c|c}
				\toprule
				\multirow{2}{*}{\bf Target} & \multirow{2}{*}{\bf Stream} & \multicolumn{4}{c}{\bf Hyper-parameters ($\lambda_{TR}$ / $\lambda_{EmbTF}$ / $\lambda_{SAL}$) for Streams of Document Collections} \\ 
				& 			& \texttt{AGnews} 			  & \texttt{TMN} 		     & \texttt{R21578} 			& \texttt{20NS} \\ \midrule
				\texttt{20NSshort}     & {\bf S1}		&     0.001 / 0.1 / 1.0  		  & 	0.001 / 0.1 / 1.0	     & 		0.001 / 0.1 / 1.0	& 0.001 / 1.0 / 1.0 \\ \hline
				\texttt{TMNtitle}        & {\bf S2}		&     0.001  / 0.1 / 0.1 		  & 	0.01 / 1.0 / 1.0	     & 		0.001 / 0.1 / 0.1	& 0.001 / 0.1 / 0.1 \\ \hline
				\texttt{R21578title}    & {\bf S3}	&     0.001 / 0.1 / 1.0  		  & 	0.001 / 0.1 / 1.0	     & 		0.001 / 1.0 / 1.0	& 0.1 / 0.1 / 1.0 \\ \bottomrule
		\end{tabular}}
	\end{center}
\end{table*}

\begin{table*}[h]
	\centering
	\caption{Label space of the document collections used}
	\label{tabsuppl:labelspace}
	\vskip 0.1in
	\resizebox{.7\textwidth}{!}{
		\begin{tabular}{c|c}
			\toprule
			{\bf data}    &    {\bf labels / classes } \\ \midrule 
			\multirow{1}{*}{TMN}    &   world, us, sport, business, sci$\_$tech, entertainment, health \\  \hline
			\multirow{1}{*}{TMNtitle}    &   world, us, sport, business, sci$\_$tech, entertainment, health \\  \hline
			\multirow{1}{*}{AGnews}    &   business, sci$\_$tech, sports, world \\  \hline

			&   misc.forsale, comp.graphics, rec.autos, comp.windows.x, \\ 
			20NS					 &   rec.sport.baseball, sci.space, rec.sport.hockey, \\ 
			20NSshort,                                 				& soc.religion.christian, rec.motorcycles, comp.sys.mac.hardware,\\
			& talk.religion.misc, sci.electronics, comp.os.ms-windows.misc,\\ 
			& sci.med, comp.sys.ibm.pc.hardware, talk.politics.mideast,\\
			&  talk.politics.guns, talk.politics.misc, alt.atheism, sci.crypt\\ \hline		
			
			&   trade, grain, crude, corn, rice, rubber, sugar, palm-oil, \\
			& veg-oil, ship, coffee, wheat, gold, acq, interest, money-fx,\\
			& carcass, livestock, oilseed, soybean, earn, bop, gas, lead, zinc,\\
			&  gnp, soy-oil, dlr, yen, nickel, groundnut, heat, sorghum, sunseed,  \\
			R21578title					& cocoa, rapeseed, cotton, money-supply, iron-steel, palladium, \\
			R21578					& platinum, strategic-metal, reserves, groundnut-oil, lin-oil, meal-feed, \\
			& sun-meal, sun-oil, hog, barley, potato, orange, soy-meal, cotton-oil,  \\
			& fuel, silver, income, wpi, tea, lei, coconut, coconut-oil, copra-cake,  \\
			& propane, instal-debt, nzdlr, housing, nkr, rye, castor-oil, palmkernel, \\ 
			& tin, copper, cpi,   pet-chem, 	 rape-oil, oat, naphtha, cpu, rand, alum 		\\ \bottomrule
	\end{tabular}}
\end{table*} 

To reproduce the scores reported, we have also provided the {\bf code} of the LNTM framework and {\bf pre-processed datasets}  used in our experiments.

\end{document}